\documentclass[journal]{IEEEtran}

\pdfoutput=1

 \usepackage{cite}

\usepackage{xcolor}

\usepackage{multirow}
\usepackage{makecell}
\ifCLASSINFOpdf
  \usepackage[pdftex]{graphicx}
\else
  \usepackage[dvips]{graphicx}
\fi
\usepackage{amsmath}
\usepackage{amssymb}

\ifCLASSOPTIONcompsoc
 \usepackage[caption=false,font=normalsize,labelfont=sf,textfont=sf]{subfig}
\else
 \usepackage[caption=false,font=footnotesize]{subfig}
\fi

\usepackage{url}
\definecolor{light-gray}{gray}{0.7}
\usepackage[switch]{lineno}

\begin{document}
%
\title{{Online Clustering-based Multi-Camera Vehicle Tracking in Scenarios with overlapping FOVs}
%
%
%

\author{Elena Luna, Juan C. SanMiguel, Jos\'{e} M. Mart\'{i}nez, and Marcos Escudero-Vi\~nolo}
\thanks{Elena Luna, Juan C. SanMiguel, Jos\'{e} M. Mart\'{i}nez, and Marcos Escudero-Vi\~nolo are with the Video Processing and Understanding Lab, Universidad Aut\'{o}noma de Madrid, Spain, e-mail: \{elena.luna;~juancarlos.sanmiguel;~josem.martinez;~marcos.escudero\}@uam.es.}

}

\markboth{}%
{Shell \MakeLowercase{\textit{et al.}}: Bare Demo of IEEEtran.cls for IEEE Journals}
%


\maketitle

\begin{abstract}
Multi-Target Multi-Camera (MTMC) vehicle tracking is an essential task of visual traffic monitoring, one of the main research fields of Intelligent Transportation Systems. Several offline approaches have been proposed to address this task; however, they are not compatible with real-world applications due to their high latency and post-processing requirements. In this paper, we present a new low-latency online approach for MTMC tracking in scenarios with partially overlapping fields of view (FOVs), such as road intersections. Firstly, the proposed approach detects vehicles at each camera. Then, the detections are merged between cameras by applying cross-camera clustering based on appearance and location. Lastly, the clusters containing different detections of the same vehicle are temporally associated to compute the tracks on a frame-by-frame basis. The experiments show promising low-latency results while addressing real-world challenges such as the a-priori unknown and time-varying number of targets and the continuous state estimation of them without performing any post-processing of the trajectories.

\end{abstract}


%
\IEEEpeerreviewmaketitle

\section{Introduction}
%
%
%
%
\IEEEPARstart{I}{ntelligent} Transportation Systems (ITS) are considered a key part of smart cities. Consistent with the accelerated development of modern sensors, new computing capabilities and communication, ITS technology engages the attention of both academia and industry. ITS point to offer smarter transportation facilities and vehicles, along with safer transport services. 

One of the main research fields on ITS is visual traffic monitoring using video analytics with data captured by visual sensors. This data can be used to provide information, such as traffic flow estimation, or to detect traffic patterns or anomalies. In recent years it has become an active field within the computer vision community \cite{Guerrero-Ibanez2018,yang2018vehicle,veres2019deep}, however it is still remains a challenging task \cite{menouar2017uav}, mainly if we consider the case of multiple cameras. 

In contrast to mono-camera traffic monitoring, multi-camera setups requires of a more complex infrastructure, the capability of dealing with more simultaneously data, as well as a higher processing capability. Multi-Target Multi-Camera (MTMC) tracking algorithms are fundamental for many ITS technologies.

Different from  Multi-Target Single-Camera (MTSC) tracking \cite{Leal-Taixe2017,Ciaparrone2020}, MTMC tracking entails the analysis of visual signals captured by multiple cameras, considering setups with overlapping fields of view (FOVs), but also scenarios for wide-area monitoring, where cameras may be separated by large distances. Road intersections are well-know targets for monitoring due to the high number of reported accidents and collisions \cite{Shirazi2017}. These intersections are known for their intrinsic and complex nature due to a variety of the vehicles' behaviors. This kind of scenarios are usually monitored with multiple partially overlapping cameras, which introduces new challenges, but also powerful opportunities for video analysis (e.g. traffic flow optimization and pedestrian behaviour analysis).

For the multi-camera tracking problem, efficient data association across cameras, and also, across frames, becomes the key problem to solve. A considerable amount of existing MTMC vehicle tracking algorithms perform an offline batch processing scheme to carry out the association  \cite{Tan2019,Chang2019,Chen2019,He2019,Hou2019,Hsu2019,Li2019,Wu2019}. They consider previous and future frames, and often the whole video sequences at once, to merge vehicles trajectories across cameras and time. They also rely on post-processing techniques to refine the resulting trajectories. This offline scheme provides more robustness, compared to online designs, albeit it is not compatible with online applications; hence, limiting its applicability in real-time traffic monitoring scenarios.

In this paper, we describe the first, to the best of our knowledge, low-latency online MTMC vehicle tracking approach for cameras with partially overlapping FOVs capturing intersection scenarios. The proposed system follows an online and frame-by-frame processing scheme. 
Furthermore, compared to other state-of-the-art systems (see Table \ref{table:comparison}), our approach does not perform any post-processing track refinement, it is agnostic to potential motion patterns (i.e. it works without prior knowledge of vehicles paths within cameras' FOV) and it does not require additional manual ad-hoc annotations (e.g. definition of regions and boundaries on the roads). These two last characteristics avoid the need of configuring each real set-up where the system is deployed, improving flexibility and generalising its use. 

The proposed MTMC tracking approach builds upon detection of multiple vehicles on every single camera. Afterwards, a combined cross-camera agglomerative clustering, combining spatial locations (using GPS coordinates) and appearance features, is used to merge vehicles from different cameras. This clustering is evaluated using validation indexes and, finally, a temporal linkage of the obtained clusters is performed to obtain the trajectories of each moving vehicle in the scene along time.

\begin{table*}
\caption{Comparison of available MTMC vehicle tracking approaches. The table shows differences regarding the type of processing, the use of post-processing tracks, the awareness about the vehicles' motion patterns, the use of ad-hoc information annotated manually and the level of cross-camera association. As can be seen, ours is the unique online approach considering detections to perform the cross-camera association, also, we do not employ any post-processing of the tracks, we are agnostic to the motion patterns of the vehicles and we do not use any additional manual annotations. \label{table:comparison} }
\begin{center}
\resizebox{0.70\textwidth}{!}{

\begin{tabular}{ccccccc}
\hline
Approach &\makecell{Processing}   & \makecell{Post-processing  \\ of tracks} & \makecell{Awareness of  \\ motion patterns } & \makecell{Ad-hoc\\ annotations} & \makecell{Level of \\ Association} \\
\hline\hline
Baidu \cite{Tan2019}                &  \multirow{8}{*}{\rotatebox{0}{Offline}}  & \checkmark    &\checkmark        &  \checkmark  &  \multirow{8}{*}{\rotatebox{0}{Tracklets}} \\
NCCU-UAlbany \cite{Chang2019}      &                             &        -       &\checkmark          &  -  & \\
CUNY NPU  \cite{Chen2019}          &                              &  \checkmark   & \checkmark        & \checkmark&\\
BUPT \cite{He2019}               &                               & -             &-                & -    &\\ 
ANU \cite{Hou2019}               &                               & -             &  -         &  -     &\\ 
UWIPL \cite{Hsu2019}              &                               & \checkmark   & \checkmark    & \checkmark &\\ 
DiDi Global \cite{Li2019}         &                             & \checkmark     & -           &   -   &\\ 
Shanghai Tech. U. \cite{Wu2019}   &                              & -              & \checkmark          &  \checkmark &\\ 
\hline
Ours                             & Online                       & -              & -              & - & Detections  \\ 
\hline
\end{tabular}}
\end{center}

\end{table*}

This paper is an extended version of our related conference publication \cite{luna2019vpulab}, with additional contributions as follows. First, we include and evaluate the impact of additional object detectors. Second, we remove any offline dependency in order to become a genuine online approach. Third, we design and train a completely new appearance feature extraction, and also investigate the impact of an additional dataset for training. Fourth, we improve the cross-camera clustering and temporal association reasoning. Fifth, we design and implement a new occlusion handling strategy. Lastly, we perform a wide ablation study to measure the impact of different parameters and strategies at different stages of the proposal, and we show results in a detailed comparison with the state-of-the-art.

The paper is organized as follows. Section \ref{sec:soa}  reviews the state-of-the-art in MTMC vehicle tracking. Section \ref{sec:method} describes the proposed approach. Section \ref{sec:results} presents the evaluation framework, the implementation details, the ablation study and finally, a comparison with the state-of-the-art. Finally, conclusion remarks are described in Section \ref{sec:conclusions}.

\section{Related work\label{sec:soa}}

For the last recent years, several approaches devoted to track pedestrians in multi-camera environments have been published \cite{Chen_2020_CVPR,He2020,Liem2014,Sio2019,zhang2019real,Zhou2020}.
The releases of public benchmarks such as MARS \cite{zheng2016mars} and DukeMTMC \cite{ristani2016performance} powered the research community to put efforts into Multi-Target Multi-Camera tracking oriented to people tracking. 

\begin{figure}
\begin{center}
    \includegraphics[width=0.49\textwidth,keepaspectratio]{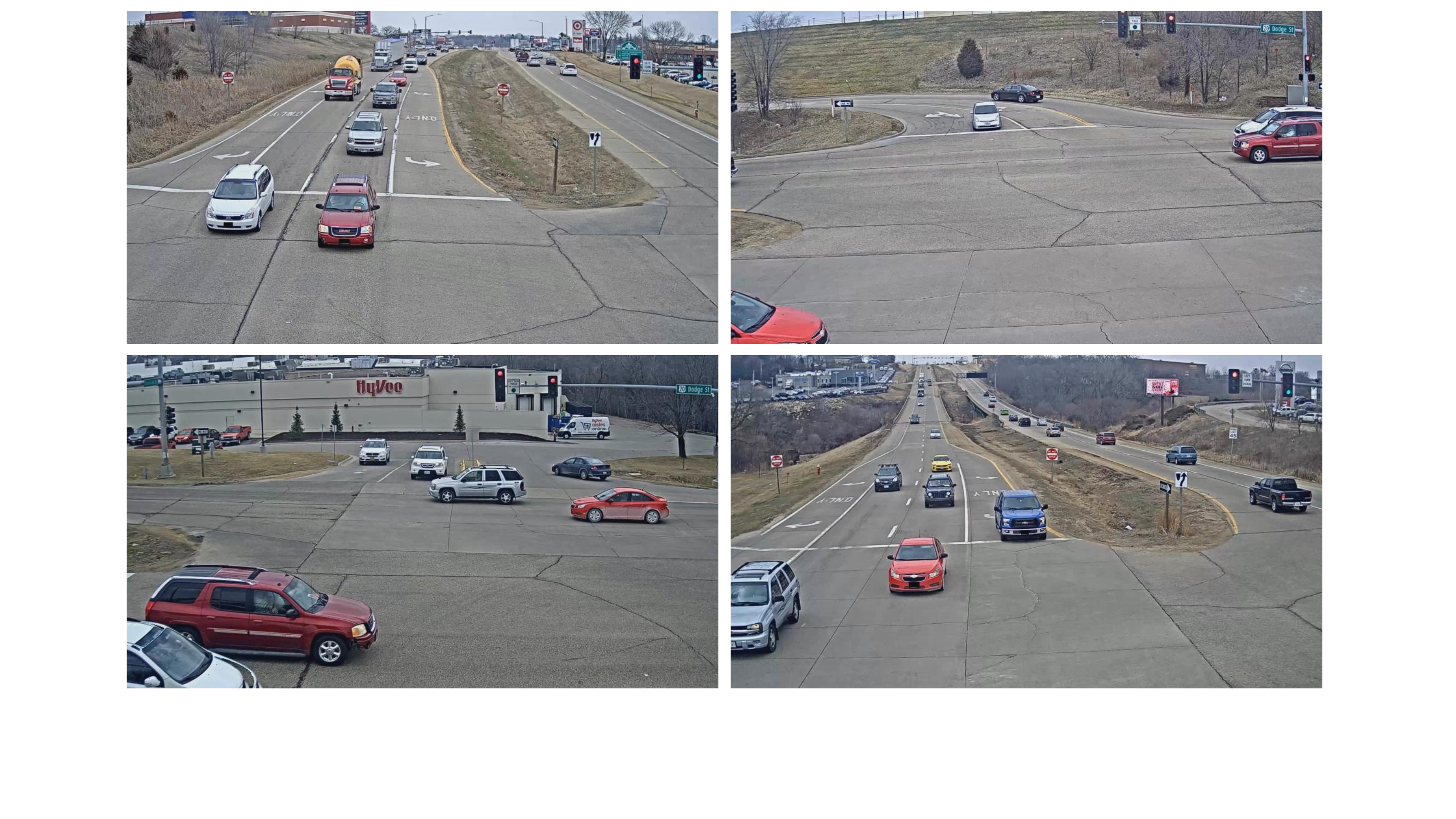}
\end{center}
   \caption{Sample views from an intersection in CityFlow benchmark. \label{fig:sample_frames}}
\end{figure}

Due to the lack of appropriate publicly available datasets, MTMC tracking focused on vehicles was a nearly unexplored field. 
To encourage research and development in ITS problems, the AI City Challenge Workshop\footnote{\url{https://www.aicitychallenge.org/}} launched three distinct but closely related tasks: 1) City-Scale Multi-Camera Vehicle Tracking, 2) City-Scale Multi-Camera Vehicle Re-Identification and 3) Traffic Anomaly Detection. Focusing on MTMC tracking, the CityFlow benchmark was presented \cite{tang2019cityflow}. At the time of publication, it is the only dataset and benchmark for MTMC vehicle tracking. Figure \ref{fig:sample_frames} depicts four sample views from an intersection in City-Flow benchmark. 

\begin{figure}
\begin{center}
    \includegraphics[width=0.49\textwidth,keepaspectratio]{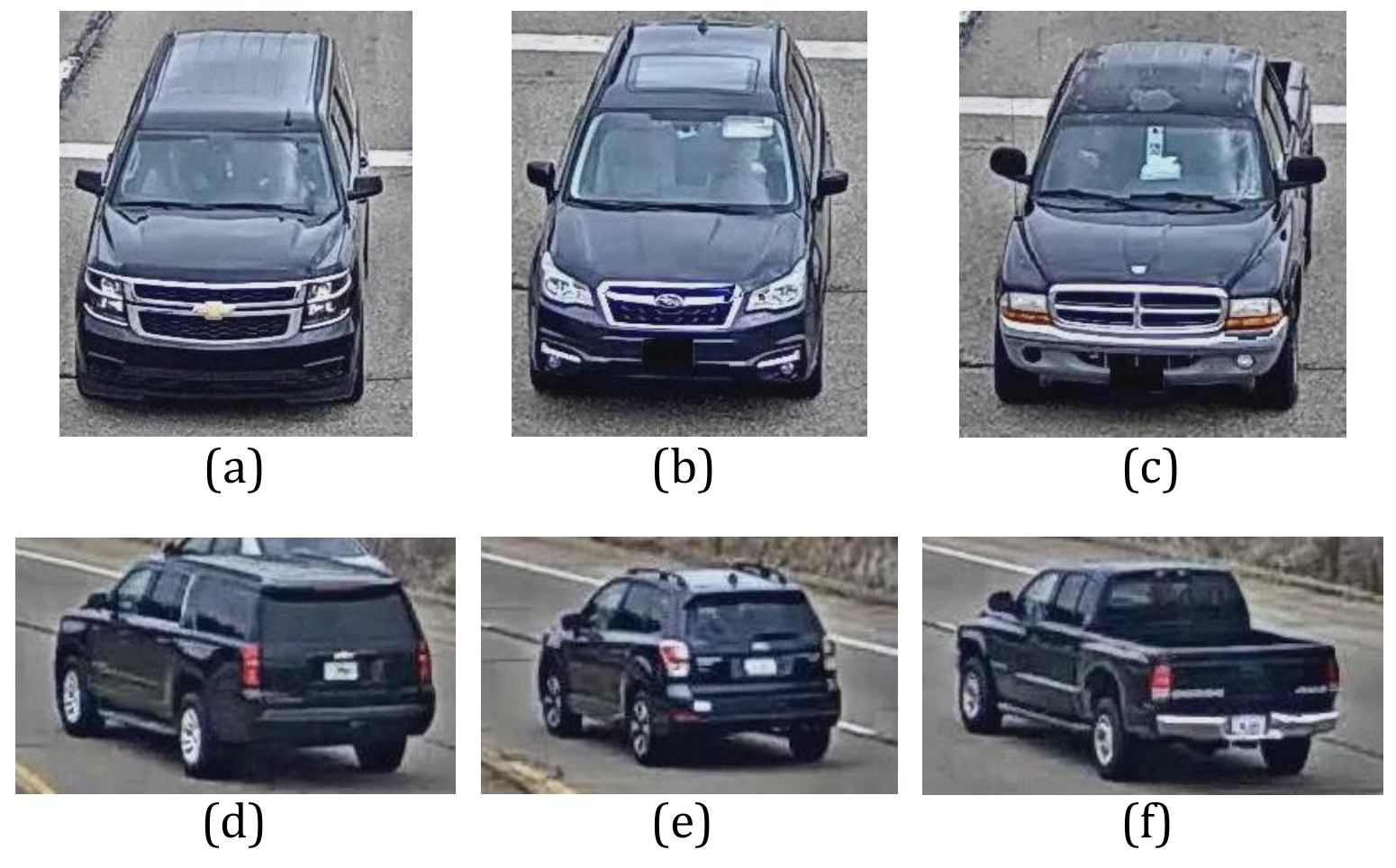}
\end{center}
   \caption{Illustration of the viewpoint variation problem. Under the same view different vehicles may appear very similar (a), (b) and (c), while the same car from different viewpoints may be extremely difficult to recognise [(a), (d)], [(b), (e)] and [(c), (f)]. \label{fig:viewpoints}}
\end{figure}

The major challenge of tracking vehicles is the viewpoint variation problem. As can be seen in Figure \ref{fig:viewpoints}, different vehicles may appear quite similar from the same viewpoint, however the same vehicle captured from different viewpoints may be difficult to recognise. It can be extremely hard, even for humans, to determine if two vehicles from different points of view depict the same car (e.g., as shown in Figure \ref{fig:viewpoints}, pairs [(a), (d)], [(b), (e)] and [(c), (f)]).

 
According to the processing scheme, MTMC tracking methods can be categorized in two groups: 1) offline methods, and 2) online methods.
Offline tracking methods, perform a  global optimization to find the optimal association using the entire video sequence. The vehicles' detections are temporally grouped into tracklets (short trajectories of detections) using MTSC tracking techniques, and, afterwards, tracklet-to-tracklet association is performed, mainly by using re-identification techniques: considering the whole video sequences at once \cite{Tan2019,He2019,Hsu2019,Li2019,Wu2019}, considering windows of frames \cite{Chen2019}, or even combining both approaches \cite{Hou2019}.

On the other side, online approaches need to perform cross-camera association of target detections on a frame-by-frame basis, using detectors' outputs (usually, bounding boxes) as the smallest unit for matching, instead of tracklets.

As can be seen in Table \ref{table:comparison}, to the best of our knowledge, all existing approaches chose to work in an offline way. 
In order to remove false positive trajectories or ID switches \cite{ristani2016performance}, the offline approaches sometimes may apply post-processing filtering at the end of some intermediate stages \cite{Tan2019,Chen2019,Li2019}, or at the end of the whole procedure \cite{Hsu2019}. 
Being aware of the motion patterns that the vehicles can adopt in every camera view, can also help to remove undesired trajectories and, therefore, increase the recall \cite{Tan2019,Chang2019, Chen2019, Hsu2019, Wu2019}. Offline working also allows to apply additional temporal constraints to increase the performance \cite{Tan2019,Hsu2019,Chen2019}.
Another strategy to improve overall performance consists in incorporating some additional manually annotated, scenario specific, information; for example, additional vehicle's attributes (colour, type, etc) for getting a better appearance model \cite{Tan2019} or road boundaries \cite{Chen2019}.

 It is common in the literature of MTMC tracking to treat the tracklet-to-tracklet cross-camera association task as a clustering problem, grouping them by appearance features \cite{ristani2018features, Tan2019, He2019}, or by combining appearance and other constraints (e.g., time and location) \cite{tang2018single,Zhang2017, Hsu2019, Chen2019}. Clustering algorithms are often categorized into two broad categories: 1) partitioning algorithms (center-based, e.g. K-means \cite{macqueen1967kmeans}, or density-based, e.g. DBSCAN \cite{ester1996density}), and 2) hierarchical clustering \cite{johnson1967hierarchical} (being agglomerative or divisive).  While hierarchical algorithms build clusters gradually (as a tree of clusters) and they do not require pre-specification of the number of clusters, partitioning algorithms learn clusters at once and they require pre-specification of the number of clusters (K-means) or the minimum number of points defining a cluster (DBSCAN). Therefore, hierarchical clustering is advantageous when there is no prior knowledge about the number of clusters, but on the contrary, it outputs a tree of clusters, commonly represented as a \emph{dendrogram}. Such structure does not provide the number of clusters, but gives information about the relations between the data. For this reason, cluster validation techniques, such as Davies-Bouldin index \cite{davies1979bouldin}, the Dunn index \cite{dunn1973fuzzy} or the Silhouette coefficient \cite{kaufman2009silh}, are used to determine the number of clusters, which may differ for each technique. In the proposed work, as there is no prior knowledge about the number of vehicles in the scene, we apply agglomerative hierarchical clustering combining location and appearance information.

 Existing MTMC vehicle tracking approaches firstly compute tracklets by temporally merging detections on every single camera, and then performing cross-camera tracklet-to-tracklet association. In contrast, we firstly compute clusters by cross-camera association of vehicle detections and, afterwards, on a frame-by-frame basis, we temporally associate the clusters to compute the tracks.

\section{Proposed Approach\label{sec:method}}

\begin{figure*}
\begin{center}
    \includegraphics[width=0.99\textwidth,keepaspectratio]{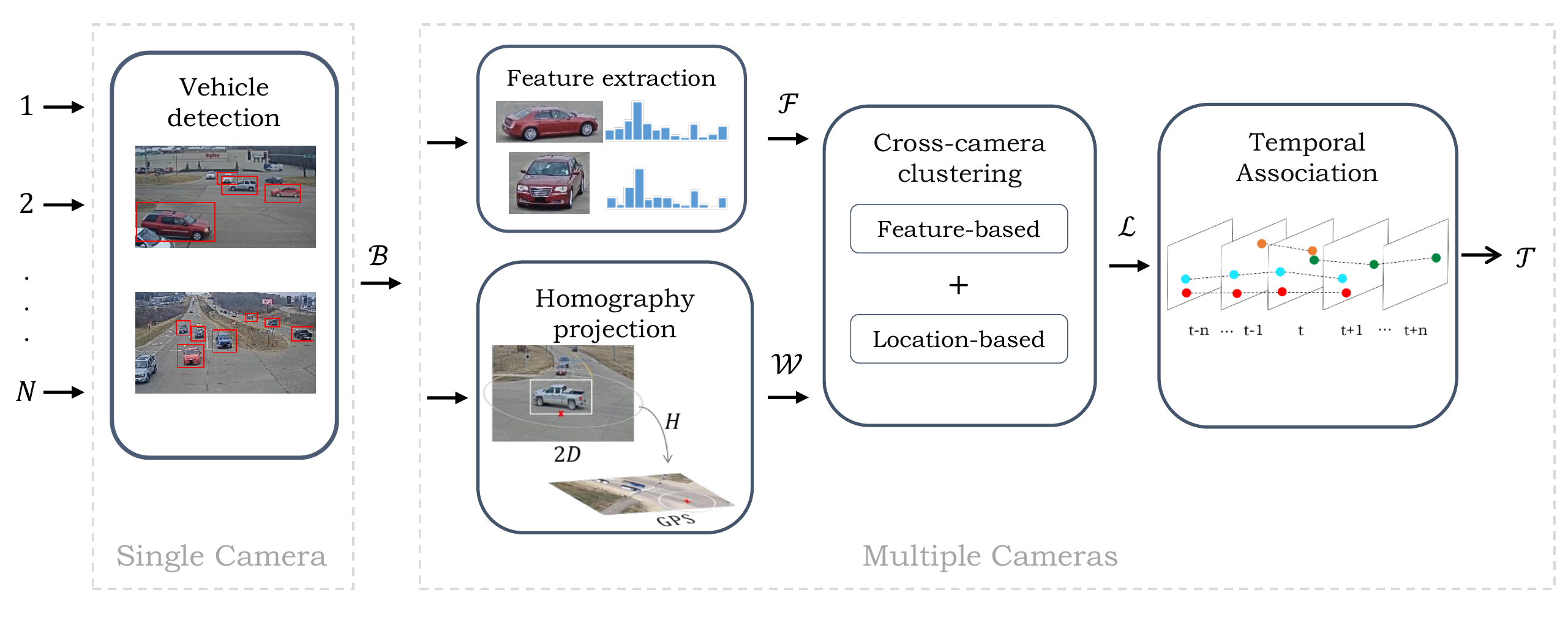}
\end{center}
   \caption{Block diagram of the proposed approach. The inputs are frames from $N$ cameras. The trajectories are computed for each frame. First, the \emph{vehicle detection} block computes $\mathcal{B}$, the set of vehicle detections. $\mathcal{B}$ feeds both \emph{feature extraction} and \emph{homography projection} blocks. $\mathcal{F}$ is the set of appearance feature descriptors and $\mathcal{W}$ the set of GPS world coordinates of every vehicle. The \emph{cross-camera clustering} block uses $\mathcal{F}$ and $\mathcal{W}$ to aggregate different views of the same vehicle and to compute the set of clusters $\mathcal{L}$ at each temporal instant.  Lastly, the \emph{temporal association} block associates clusters in $\mathcal{L}$ in a temporal way to compute the set of tracks $\mathcal{T}$.  \label{fig:block_diagram}}
\end{figure*}

In the proposed online Multi Target Multi Camera tracking approach, all cameras' videos are processed simultaneously frame by frame, without any post-processing of the trajectories. The approach is composed of five processing blocks, as shown in Figure~\ref{fig:block_diagram}. 
As input, we consider a network of calibrated and synchronized cameras with partially overlapping FOVs providing independent video sequences. Given a network of $N$ cameras, the pipeline includes the following stages: (1) vehicle detection, (2) feature extraction, (3) homography projection, which projects single camera vehicles from each camera to the world coordinates system (GPS) for providing location information, (4) cross-camera clustering, that is fed on the output of (2) and (3) blocks, and (5) temporal association of vehicles trajectories over time to compute the tracks. As result, the system generates tracks consisting on the identity and location of every vehicle along time. The design of the processing blocks is detailed in the following subsections, whilst the implementation details are given in Section \ref{sec:implementation-detail}.

Table \ref{tab: notation} summarizes the notation used in this section. The scope of each variable is also defined. \emph{Scenario} refers to the set of cameras, \emph{Frames} stand for all the simultaneous images coming from the cameras at each temporal instant. \emph{Sequence} is comprised of all the aggregated frames coming simultaneously from the cameras. $N$, and $\textbf{H}_n$  are intrinsic to the scenario, while $r$ is a design parameter. $D$, $\mathcal{B}$, $\mathcal{W}$, $\mathcal{F}$, and $\mathcal{L}$ are computed at each temporal instant, needing the simultaneous frames. Last, $\mathcal{T}$ is updated frame-by-frame for the whole sequence.

\begin{table}
\caption{Notation used throughout the paper. \label{tab: notation}}
\begin{center}
\resizebox{0.47\textwidth}{!}{
\begin{tabular}{ccc}
\hline
\makecell{Symbol } & \makecell{Description } & \makecell{Scope} \\
\hline\hline
$N$ & Number of cameras & Scenario \\

$\textbf{H}_n$ & Homography matrix of $n^{th}$ camera  & Scenario \\
$r$ & Association radius & Scenario\\
$D$ & Number of total detections & Frames\\
$\mathcal{B}$  & Set of bounding boxes & Frames\\
$\mathcal{W}$ & Set of GPS world locations  & Frames\\
$\mathcal{F}$ &  Set of feature descriptors & Frames\\
$\mathcal{L}$ & Set of clusters  & Frames\\
$\mathcal{T}$ & Set of tracks  & Frames, Sequence\\
\hline
\end{tabular}}
\end{center}

\end{table}

\subsection{Vehicle Detection} \label{sec: prop-vehicledet}

As most of the state-of-the-art MTMC tracking methods, we follow the tracking-by-detection paradigm. Therefore, the first stage of the pipeline is vehicle detection at each frame. Let $\textbf{b}=[x, y, w, h]$ be a bounding box with $[x,y]$ being the upper-left corner pixel coordinates, and $[w,h]$ the width and height. Let define $\mathcal{B}=\left\{\textbf{b}_{d}, d \in[1, D]\right\}$ as the set of bounding boxes at each frame for all the cameras, with $D$ the total number of detections. 

Note that the proposal can incorporate any single-camera vehicle detection algorithm whose output is in a bounding box form. 

\subsection{Feature Extraction}\label{sec:appearance-features}
In order to describe the appearance of the $d^{th}$ bounding box detection, let $\textbf{f}_{d}$ be its $k$-dimensional deep feature descriptor. Let $\mathcal{F}=\left\{\textbf{f}_{d},d \in[1, D]\right\}$ be the set of appearance feature descriptors for each frame for all the detected vehicles.

Due to the intrinsic geometry of vehicles, their appearance may suffer strong variations across different camera views. This variance is such that it could be, even for a human being, very hard to determine if they are the same vehicle. Thus, in order to have highly discriminating features, we trained a model to improve vehicle classification ability in the faced scenario. More details on this vehicle specific model will be given in Section \ref{sec: imp-feature}.

Class imbalance is a form of the imbalance problem \cite{Oksuz2020}, that occurs when there is an important inequality regarding the number of examples pertaining to each class in the data. When not addressed, it may have negative effects on the final performance. 
It is known that classes with a higher number of observations tend to dominate the learning process, hindering the learning and generalization of low-represented classes. In order to minimize the imbalance effects, instead of classical Cross-Entropy (CE) loss  \cite{goodfellow2016deep}, we employ the focal loss (FL) proposed in \cite{Lin_2017_ICCV}.


\subsection{Homography-based Projection}\label{sec:projection}

This processing block computes the location of each detected vehicle on the common ground-plane employing GPS coordinates. Let $\textbf{H}_n$ be the homography matrix that transforms coordinates from the image plane of the $n^{th}$ camera to the GPS coordinates of the common ground plane. Let the inverse matrix $\textbf{H}_n^{-1}$ be the inverse transformation.
We leverage the GPS coordinates to achieve a high-precision clustering based on the location information by applying camera projection. Given a bounding box $\textbf{b}$, one can obtain its associated GPS  coordinates, i.e. $[\phi, \lambda]$ (latitude and longitude), by projecting the middle point of its base with the $\textbf{H}_n$ transformation. $\mathcal{W}=\left\{[\phi, \lambda]_{d}, d \in[1, D]\right\}$, the set of GPS coordinates, is obtained after applying the transformation to the set $\mathcal{B}$.
 
\begin{figure}
\begin{center}
    \includegraphics[width=0.49\textwidth,keepaspectratio]{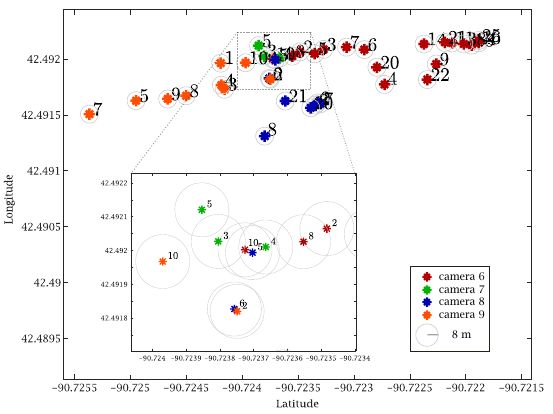}
\end{center}
   \caption{Vehicle detections from four partially overlapped cameras projected to GPS coordinates at a certain temporal instant. Detections within a 8 meters radius are more likely to be joined. (Best viewed in color)   \label{fig:projections}}
\end{figure}

Figure \ref{fig:projections} illustrates an example of the projected detections coming from different cameras.
Note that this block relies on the output of the object detection stage, and along with the feature extraction module, it feeds the cross-camera clustering.

\subsection{Cross-camera Clustering}\label{sec:proposed-clustering}

Given the sets $\mathcal{B}$,  $\mathcal{W}$ and $\mathcal{F}$, the cross-camera clustering block associates different camera views of the same vehicle at each frame to compute $\mathcal{L}=\left\{l_{i}, i \in[1, L]\right\}$, the set of clusters at a given frame, being $L$ be the number of created clusters. Clusters' content ranges from a single detection, if the vehicle is only visible by one camera, to the maximum number of detections, defined by the maximum number of cameras capturing the scene. To create the clusters, we compute a frame-by-frame linkage by performing an agglomerative hierarchical clustering combining location and appearance features. 

Hierarchical clustering \cite{johnson1967hierarchical} requires a square \emph{connectivity} matrix of distances (dissimilarities) or similarities of the input data to merge. We compute the \emph{connectivity} matrix $\Theta$ as a constrained pairwise features distance between all the vehicles coming from every camera at each frame. 
At each frame, we compute the pairwise Euclidean distance between the appearance feature vectors of all the vehicles under consideration, as follows:
\begin{equation}
 \zeta _{d,d'} = ||\textbf{f}_{d}-\textbf{f}_{d'}||_{2}
\end{equation}

Also at each frame, we compute the Euclidean pairwise distance between all the GPS coordinates of vehicles:
\begin{equation}
  \psi _{d,d'} = ||(\phi_{d}-\phi_{d'})^{2}-(\lambda_{d}-\lambda_{d'})^{2}||_{2}
\end{equation}

The spatial distance and the camera ID are used to apply some constraints. Since two vehicles' detections widely separated in GPS coordinates are highly unlikely to come from the same vehicle, it is reasonable to assume a maximum association distance. This constraint narrows down the list of vehicles to be matched and improves the ability to distinguish different identities by focusing on comparing only nearby targets. Hence, the connectivity matrix $\Theta $ is computed as follows:
\begin{equation}
\Theta'_{d,d'}=\left\{\begin{array}{ll}
\zeta_{d,d'}, & \psi_{d,d'} \leq r \\
\infty, & \psi_{d,d'} > r
\end{array}\right.,
\end{equation} being $r$ the maximum association radius.
A second condition is applied for preventing vehicles' detections from the same camera view to be merged together. It is done by constraining the association matrix as follows:
\begin{equation}
\Theta_{d,d'}=\left\{\begin{array}{ll}
\Theta'_{d,d'}, &  c_d\neq c_{d'} \\
\infty, & c_d= c_{d'}
\end{array}\right.
\end{equation}
Let $c_{d}$ be the camera yielding the $d^{th}$ detection.

\begin{figure}
\begin{center}
    \includegraphics[width=0.49\textwidth,keepaspectratio]{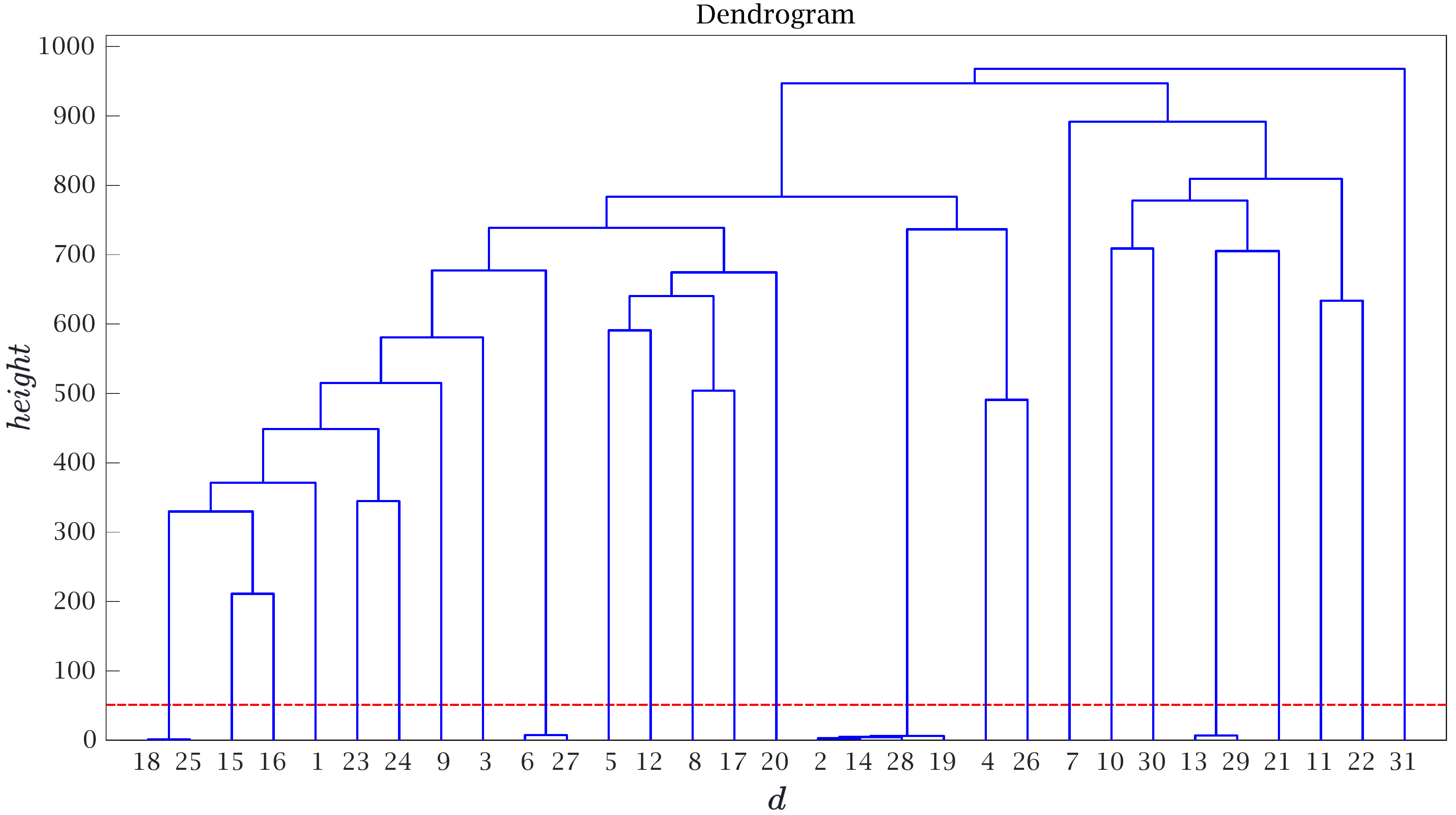}
\end{center}
   \caption{Dendrogram illustrating the hierarchical relationships between all the detected vehicles $d \in [1,29]$ at a certain temporal instant. After cluster validation, the dendrogram is cut at $height=50$: detections that are joined together below the red line are part of the same cluster. In this example (18, 25), (6, 27), (2, 14, 28, 19) and (13, 29) are joined together, while the rest of detections comprise a cluster by themselves. Hence, $L=25$.  \label{fig:dendogram}}
\end{figure}

As stated above, hierarchical clustering methods departs from a connectivity matrix $\Theta$ to compute a tree of clusters and this structure does not provide the number of clusters, but gives information on the relations between the data. These relationships can be represented by a tree diagram called dendrogram. An example is presented in Figure \ref{fig:dendogram}. In order to cut the dendrogram and identify the optimal number of clusters, we use the Dunn index \cite{dunn1973fuzzy} for cluster validation. The aim of this index is to find clusters that are compact, with a small variance between members of the cluster, and well separated by comparing the minimal cluster distance to the maximal cluster diameter. The cluster diameter is defined as the distance between the two farthest elements in the cluster. This process provides the number of vehicles at every frame in the scene, in the form of clusters, as well as its location, in the form of the cluster's centroid (outlined as the mean point at each coordinate axis of all the components). To sum, at every frame, each cluster designates an existing vehicle. 

\subsection{Temporal Association}\label{sec:prop-temporal}

The last stage of the proposed approach links clusters over time to estimate the vehicle tracks. Let $\textbf{t}_j = [\textbf{x}^{s_{start}}_{j},...,\textbf{x}^{s_{end}}_{j}]$ be the $j^{th}$ track defining the trajectory of a moving vehicle by a succession of states. Each state is described by $\textbf{x}^{s}_{j}=[\phi,\lambda,v_{\phi},v_{\lambda}],$ where $[\phi, \lambda]$ is the target location and $[v_{\phi},v_{\lambda}]$ is the target velocity, both represented using GPS coordinates. Let define $\mathcal{T}=\left\{\textbf{t}_{j}, j \in[1, J]\right\}$ as the set of tracks along the video sequence. In contrast to previous sets $\mathcal{B}$, $\mathcal{W}$ and $\mathcal{F}$, that are initialized at each frame, $\mathcal{T}$ is built incrementally, i.e. it is computed at first frame and updated along time. In other words, tracks depict the location of clusters along time. As in the whole system, the temporal association is performed on-line, that is, frame-by-frame.

Vehicles' motion is estimated using a constant-velocity Kalman filter~\cite{kalman1960new}. The Kalman filter makes a prediction of the state of the target as a combination of the targets' previous state (at prior frame) and the new measurement (at current frame) using a weighted average. It results in a new state estimation lying in between the previous prediction and the measurement. Thus, at each frame, on the one hand we employ Kalman filter to get the estimate location of the tracks of the previous frame, and on the other hand, we get the current vehicle measurements as the clusters resulting from the cross-camera association. 

In order to associate both, we apply the Hungarian Algorithm~\cite{kuhn1955hungarian} to solve the assignment problem, using an association matrix to enumerate all possible assignments. The association matrix is filled with the pairwise L2-norm, i.e. the euclidean distance, between the location of the estimated tracks and the clusters' centroid location (see Section \ref{sec:proposed-clustering}).

To provide robustness against occlusions we designed two strategies: a blind occlusion handling and a reprojection-based occlusion handling. The first maintains alive the tracks during a short time when the detections associated to it are lost. Keeping on predicting the position of the track during that period allows to recover it in case the detections are recovered. This is helpful if the vehicle detector loses a detection, either due to a bad detection performance or a hard occlusion. The second strategy detects if a track has lost one or more of its associated detections and looks for the same track in the previous frame to get the information about the size of its previously associated bounding boxes. The new location in the current frame is inferred by applying the corresponding inverse homography matrix (e.g. $\textbf{H}_{n}^{-1}$ assuming a detection is missing for the $n^{th}$ camera) to the estimated track position. Therefore, when this strategy reveals a track which detection or detections are lost, mostly due to an occlusion the detector cannot deal with, we can generate an artificial detection with accurate estimates on the correct position and with the previous detected size of the occluded vehicle.

\section{Experiments\label{sec:results}}

\subsection{Evaluation framework}
\subsubsection{Datasets}
We considered the CityFlow benchmark \cite{tang2019cityflow}, since there is no other publicly available dataset devoted to MTMC vehicle tracking with partially overlapping FOVs. The  dataset comprises videos of 40 cameras, 195 total minutes recorded for all cameras, and manually annotated ground-truth consisting of 229,690 bounding boxes for 666 vehicles. The dataset is divided into 5 scenarios (S01, S02, S03, S04 and S05) covering intersections and stretches of roadways. S01 and S02 have overlapping FOVs, while S03, S04 and S05 are wide-area scenarios. The CityFlow benchmark also provides the camera homography matrices between the 2D image plane and the ground plane defined by GPS coordinates based on the flat-earth approximation.

We have also used VeRi-776 dataset for improving the feature extraction model by using it as additional training data. VeRi-776 \cite{shen2017veri}  is one of the largest and most common datasets for vehicle re-identification in multi-camera scenarios. It comprises about 50,000 bounding boxes of 776 vehicles captured by 20 cameras. 

\subsubsection{Evaluation Metrics\label{sec: evaluation-metrics}}
The MTMC tracking ground-truth provided by the CityFlow benchmark consists of the bounding boxes of multi-camera vehicles labeled with consistent IDs. 

Following the CityFlow benchmark evaluation methodology, Identification Precision, Identification Recall and $F_1$ Score measures \cite{ristani2016performance} are adopted:
\begin{equation}
IDP=\frac{IDTP}{IDTP+IDFP} \,  ,
\end{equation}
\begin{equation}
IDR=\frac{IDTP}{IDTP+IDFN}\, ,
\end{equation}
\begin{equation}
IDF_{1}=\frac{2 \cdot IDTP}{2\cdot IDTP+IDFP+IDFN} \, ,
\end{equation}
where $IDTP$, $IDFP$ and $IDFN$ stand for True Positive ID, False Positive ID and False Negative ID, respectively. $IDP$ ($IDR$) is the fraction of computed (ground-truth) tracks that are correctly identified. $IDF_1$ is the ratio of correctly identified tracks over the average number of ground-truth and computed tracks.

Automatically obtained tracks by the proposed method are pairwise compared with the ground-truth tracks. We declare a match, i.e. an $IDTP$, when two tracks temporarily coexist and the area of the intersection of the bounding boxes is higher than $\tau_{IoU}$ (with $0 < \tau_{IoU} < 1$) times the area of the union of the two boxes. Hence, $\tau_{IoU}$ is the Intersection Over Union (IoU) threshold.
A high $IDF_1$ score is obtained when the correct multi-camera vehicles are detected, accurately tracked within each camera view, and labeled with a consistent ID across all the views in the dataset.

\subsubsection{Hardware and software}
The algorithm and model training have been implemented using PyTorch 1.0.1 Deep Learning framework running on a computer using a 6 Cores CPU and a NVIDIA GeForce GTX 1080 12GB Graphics Processing Unit. 

\subsection{Implementation details\label{sec:implementation-detail}}

\subsubsection{Vehicle detection} \label{sec: imp-vehicle}
Regarding single-camera vehicle detection we have experimented with public detections, i.e. vehicle detections provided by the CityFlow Benchmark, and private detections, computed using a state-of-the-art algorithm.
The public detections were obtained by using three popular detectors: Yolo v3~\cite{redmon2018yolov3}, SSD512~\cite{liu2016ssd} and Mask R-CNN \cite{he2017mask}. Yolo v3 is a one-stage object detector that solves detection as a regression problem. SSD512 is also a single-shot detector which directly predicts category scores and box offsets for a fixed set of default bounding boxes of different scales at each location. Mask R-CNN, on the contrary, is a two-stage detector consisting of a region proposal network that feeds region proposals into a classifier and a regressor.  

Moreover, we have complemented the provided detections with those obtained by the EfficientDet \cite{tan2020efficientdet} algorithm, a  top-performing state-of-the-art object detector. EfficientDet is also a one-stage detector that uses EfficientNet \cite{tan2019efficientnet} as the backbone network and a bi-directional feature pyramid feature network (BiFPN).

All these approaches make use of pre-trained models on the COCO  benchmark ~\cite{lin2014microsoft}. For our purpose, we considered only detections classified as instances of the car, truck and bus classes.

\subsubsection{Feature extraction} \label{sec: imp-feature}
For the feature extraction network, we employ ResNet-50 \cite{he2016deep} as backbone, but the original classification layer (\emph{fc\_1} layer), shaped for image classification on the ImageNet dataset \cite{deng2009imagenet}, is replaced by a new classification layer whose size is tailored to the total number of identities in the training data. In order to leverage the pretained weigths on Imagenet, we fine-tune the network but freeze it until \emph{conv\_5} layer. 

To fine-tune the network, we used the CityFlow benchmark training data (S01, S03 and S04) and we also included the VeRi-776 dataset, bringing a total of 905 vehicle IDs for training (129 IDs from CityFlow, plus 776 IDs from VeRi-776). Since only training identities are known, the network learns features to correctly classify the 905 different training vehicle identities. We perform a validation methodology on pairs of unseen vehicles and comparing whether predictions are the same or not. Therefore, we check the network ability to discern different views of the same target. To create these pairs, we randomly select half of the data from S05 scenario to create a 169 IDs validation set. We forced the validation batch to contain approximately 50$\%$ of positive and 50$\%$ of negatives pairs. The pair selection is randomly done over the set of IDs, instead of the set of images, thus, IDs containing few samples are not impaired. At inference, we adopt, as a 2048-dimensional descriptor, the output of the average pooling layer, just before the classifier. 

Each input image containing a bounding box of a vehicle is adapted to the network by resizing it to $224$x$224$x$3$ and pixels' values are normalized by the mean and standard deviation of the ImageNet dataset. In order to reduce model overfitting and to improve generalization, we perform several random data augmentation techniques such as horizontal flip, dropout, Gaussian blur and contrast perturbation.

To minimize the loss function and optimize the network parameters, we adopt Stochastic Gradient Descend (SGD) solver. Experimentally, the initial learning rate was set to 0.1 and we follow a step decay schedule dropping it by 0.1 every 25 epochs. Momentum was set to 0.9 and weight decay to $1e^{-4}$.

\subsection{Ablation Study\label{sec:initial tests}}

This section measures the impact of the strategies used along the different stages of the proposed approach. Firstly, the effect of using different vehicle detectors is evaluated. Secondly, the influence of the association radius parameter is analysed. Subsequently, we gauge the influence of the appearance model training method as well as the size of the feature embedding. And finally, some additional strategies (e.g. occlusions handling) are assesed.
All the experiments are evaluated on the testing scenario of the CityFlow benchmark dataset with partially overlapping FOVs, i.e. the S02 scenario. It is composed of 4 cameras pointing to an intersection roadway (see Figure \ref{fig:sample_frames}). In total, aggregating 129 annotated vehicles whose trajectories are distributed along 8440 frames (2110 per camera) that have been captured at 10 fps.

\begin{table}
\caption{\label{tab:detectors}MTMC tracking performance of the proposed approach for different vehicle detectors. We differentiate between the public detectors (SSD512, Yolo v3 and Mask R-CNN) and  and the private one (EfficientDet). For each detector, we include the mean Average Precision (mAP) for object detection task in COCO dataset \cite{lin2014microsoft} as a measure of performance. Best of both categories in bold. \label{tab: vehicle-detector}}
\begin{center}
\resizebox{0.49\textwidth}{!}{
\begin{tabular}{cccccc}
\hline
 \makecell{COCO\\ mAP } &\makecell{Vehicle \\ Detector } & \makecell{Score \\ Threshold } &\makecell{$IDP\uparrow$}  & \makecell{$IDR\uparrow$} &\makecell{$IDF_1\uparrow$}   \\
\hline\hline
 \multirow{3}{*}{28.8} &\multirow{3}{*}{SSD512}    & 0.1 & 48.96 & 63.43 &55.27\\
 &                           & 0.2 & 48.97 &63.43 & 55.27\\ 
 &                           & 0.3 & 49.05 & 60.87 & 54.33\\ 
  \multirow{3}{*}{33.0} &\multirow{3}{*}{Yolo v3}    & 0.1  & 41.16 & 60.54 & 49.00 \\ 
&                            & 0.2  & 41.06 & 60.37 & 48.88 \\ 
&                            & 0.3 & 39.83 & 55.25 & 46.23  \\ 
                         
 \multirow{3}{*}{40.3}&\multirow{3}{*}{Mask R-CNN} & 0.1 & 49.11 &64.84 & 55.89\\  
&                            & 0.2 & \textbf{49.11} & \textbf{64.84} & \textbf{55.89} \\
&                           & 0.3 & 48.27 & 63.63 & 54.89 \\ 
   \hline
 \multirow{3}{*}{55.1}&\multirow{3}{*}{EfficientDet} & 0.1  & 39.93 & 60.85 & 48.22\\ 
 &                           & 0.2  & 44.35 & 65.12 & 52.76 \\ 
  &                          & 0.3 & \textbf{47.10} & \textbf{68.21} & \textbf{55.72} \\

\hline
\end{tabular}}
\end{center}

\end{table}

\textbf{Influence of the vehicle detector algorithm:} Table \ref{tab: vehicle-detector} comprises the impact of different vehicle detectors on the overall performance of the proposed approach. As stated before, we consider three provided object detections coming from Yolo v3, SSD512 and Mask R-CNN, i.e. public detections. We also evaluate the performance of EfficientDet, a top-performing algorithm.

We experimented with three different score thresholds to get the output detections (0.1, 0.2 and 0.3). Regarding the public detections, one can observe that the compared detectors achieve the peak performance when a low threshold is applied. The results suggest that filtering the output detections by scores higher than 0.2 leads to a lower IDR in the MTMC tracking performance. This finding indicates that detections with low confidence (mostly generated by remote and partially visible vehicles) are still useful. 

On the contrary, EfficientDet, since it is a better performing objecter detector, results in a higher $IDR$ and $IDP$ being filtered with 0.3 instead of 0.2. It enhances $IDR$ by 3.37, compared with the best results of Mask-RCNN, however, $IDP$ is degraded by 2.01. The reason for this decline is that EfficientDet is providing more False Positive trajectories arising from the detection of partially occluded vehicles that Mask-RCNN is not able to detect.

In the light of these results, 
we opted for adopting Mask R-CNN output detections filtered by 0.2 score as public detections, and EfficientDet output filtered out by 0.3, as private detections for the rest of the experiments.

\begin{table}
\caption{\label{tab:radius}Impact of the association parameter $r$ over the MTMC tracking performance. Best in bold. }
\begin{center}
\resizebox{0.49\textwidth}{!}{
\begin{tabular}{ccccc}
\hline
&Association radius  &\makecell{$IDP\uparrow$}  & \makecell{$IDR\uparrow$} &\makecell{$IDF_1\uparrow$}   \\
\hline\hline

\multirow{4}{*}{\rotatebox{90}{\tiny Mask R-CNN}}  &$r$ = 5 m. & 47.04 & 62.11 & 53.53 \\ 
&$r$ = 6.5 m. & 46.03 & 60.78 & 52
39\\
&$r$ = 8 m &  \textbf{49.11} & \textbf{64.84} & \textbf{55.89}\\
&$r$ = 9.5 m. & 46.56 & 61.47 & 52.99    \\
\hline
\multirow{4}{*}{\rotatebox{90}{\tiny EfficientDet}}  &$r$ = 5 m. & \textbf{47.18} & \textbf{68.37} & \textbf{55.83}\\ 
&$r$ = 6.5 m. & 46.41 & 67.24 & 54.92 \\
&$r$ = 8 m. & 47.10 & 68.21 & 55.72 \\
&$r$ = 9.5 m. &  43.24&62.59 &61.14   \\
\hline
\end{tabular}}

\end{center}
\end{table}

\textbf{Influence of the association radius:} Table \ref{tab:radius} shows how the association radius $r$, used in the cross-camera clustering (see Section \ref{sec:proposed-clustering}), affects the MTMC tracking performance of the proposed approach in the evaluated scenario. We sweep radius values of 5, 6.5, 8 and 9.5 meters. The results on the table indicate that the choice of the radius is quite relevant, having a significant impact in the performance, and also it is highly-dependant on the detection algorithm.
The Mask-RCNN detector gets performance peak for $r=8$, however, when using the EfficientDet detector a smaller radius, $r=5$, is the optimal choice. The reason of this difference may be related with bounding box accuracy (i.e. how the output bounding boxes fit the vehicles). Since the middle points of the bases of the bounding boxes from different camera views are projected to the ground-plane, the tighter the boxes are, the more accurate are the projections. 

Due to the common vehicles dimensions, it may be natural to think that a smaller radius should be enough to successfully associate several detections of the same vehicle. However, due to noise in the video transmission while capturing the data, some frames are skipped within some videos, so some cameras suffer from a subtle temporal misalignment (i.e. they are unsynchronized with respect to the others). Therefore, the optimal $r$ values for the CityFlow benchmark using the proposed algorithm are 5 and 8 meters, given the two evaluated detectors. 


\textbf{Influence of the appearance feature model:}  Table \ref{tab: feature-model} summarizes the effect of the training schemes on the  model used to describe the appearance features of vehicles for the proposed MTMC tracking approach. The table lists the data that has been used for training the network (described in Section \ref{sec: imp-feature}) and how the weights of the network were obtained. As the baseline, we use the model pretrained on the Imagenet dataset. As training data, we considered the training set of the CityFlow benchmark (S01, S03 and S04 scenarios) and also the training set of the CityFlow benchmark jointly with the VeRi-776 dataset. We tried two classification loss functions: Cross-Entropy loss (CE loss) and Focal Loss (FL).

Table \ref{tab:feature model} indicates that the tracking performance behaviours in a coordinated manner using both Mask R-CNN and EfficientDet detectors. In both cases, fine-tuning the network to the CityFlow benchmark has a slightly, but positive, influence. Including more training data, the VeRi-776 dataset, appears to improve the quality of the feature embeddings, resulting in a even better tracking performance.

\begin{figure}
\begin{center}
    \includegraphics[width=0.49\textwidth,keepaspectratio]{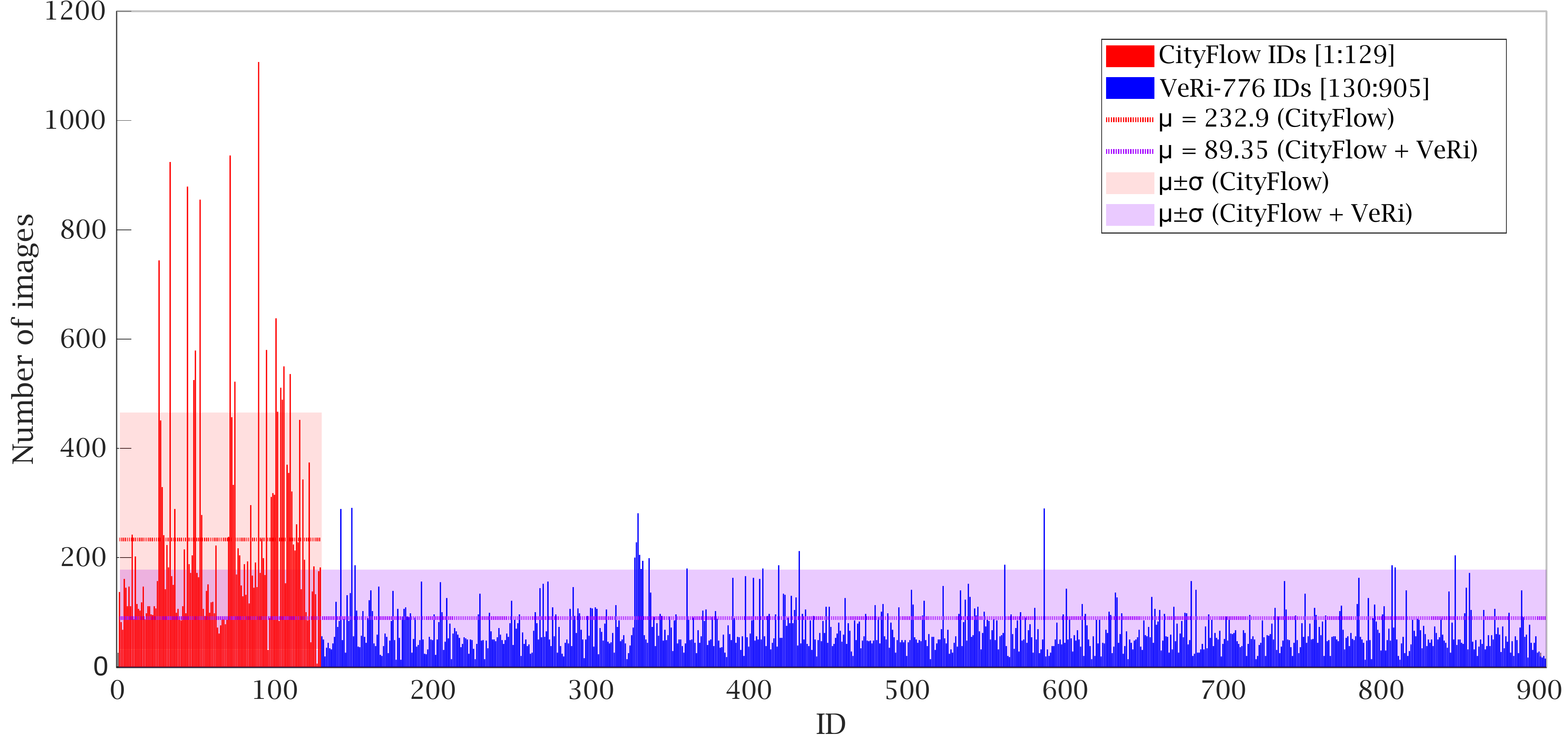}
\end{center}
   \caption{The distribution of the number of images per vehicle identity in the CityFlow training dataset, the VeRi-776 datastet and the distribution of both joined. Best viewed in color. \label{fig:distribution}}
\end{figure}

Figure \ref{fig:distribution} depicts in red the distribution of the number of images per vehicle ID of the training set of the CityFlow dataset illustrating that it is a quite unbalanced set with a very scattered distribution.  The average of the distribution is $\mu_{city} = 232.90$, while the standard deviation is $\sigma_{city} = 201.19$. From Table \ref{tab: feature-model}, we observe that training the CityFlow benchmark with the Focal loss, instead of the Cross-Entropy loss, has a positive influence in our MTMC tracking approach. 

Figure \ref{fig:distribution} also depicts in blue the distribution of the number of images per vehicle ID of the VeRi-776 dataset, as one may observe, it is more balanced than the CityFlow set. Considering both datasets together, the join distribution is now described by $\mu_{join} = 89.35$ and $\sigma_{city} = 102.25$, as $\sigma_{join} << \sigma_{city}$, one could say that the join dataset is less disperse than the single CityFlow, which can be an indicator for the subtle increase in performance obtained when the combined dataset is used.  According to these results, we opt for using the combined dataset and the CE loss for the rest of the experiments.

\begin{table}
\caption{\label{tab:feature model}  Impact of appearance feature model over the MTMC tracking performance. F: Finetuned. CE: Cross-Entropy Loss. FL: Focal Loss. Best in bold. \label{tab: feature-model} }
\begin{center}
\resizebox{0.49\textwidth}{!}{
\begin{tabular}{cccccc}
\hline&\makecell{Training  Data} & Weights  &\makecell{$IDP\uparrow$}  & \makecell{$IDR\uparrow$} &\makecell{$IDF_1\uparrow$}  \\
\hline \hline
\multirow{5}{*}{\rotatebox{90}{Mask R-CNN}} &Imagenet & Pretrained  &  49.11 & 64.84 & 55.89  \\
\cline{2-6}
&\multirow{2}{*}{CityFlow}&F +  CE  & 49.16 & 64.91 & 55.95 \\
&                        &F + FL  &49.26 & 65.04 & 56.06 \\ 
\cline{2-6}
&\multirow{2}{*}{CityFlow + VeRi-776} &F + CE & \textbf{50.56} &\textbf{66.75} & \textbf{57.54}  \\
&                        &F + FL  & 49.53 & 65.39 & 56.37 \\
\hline
\multirow{5}{*}{\rotatebox{90}{EfficientDet}} &Imagenet & Pretrained  &  47.18 & 68.37 & 55.83 \\
\cline{2-6}
&\multirow{2}{*}{CityFlow}&F +  CE  & 47.41 & 68.70 & 56.11 \\
&                         &F + FL   & 47.43 & 68.73 & 56.13 \\ 
\cline{2-6}
&\multirow{2}{*}{CityFlow + VeRi-776} &F + CE & \textbf{48.33} &\textbf{70.03} & \textbf{57.19}  \\
&                        &F + FL  & 47.63 & 69.01 & 56.36\\
\hline
\end{tabular}}

\end{center}
\end{table}

\textbf{Influence of size of the feature embedding:} Table \ref{tab:embedding-size} comprises the experiments carried out to explore the effect of the size of the feature embeddings. As stated in Section \ref{sec: imp-feature}, the output of the last average pooling layer of ResNet-50 provides a 2048-dimensional embedding. We set this embedding size as the baseline. In order to modify the length of the embedding, an additional fully connected layer of size 512, 1024 or 4096 is added at the end of the network. The additional fully connected layer is preceded by batch normalization and ReLU layers, and the training procedure is the same as described in Section \ref{sec: imp-feature}. The performance suggests that adding an additional layer, and therefore, more complexity to the model, either to reduce or increase the embedding size, may harm the performance, leading the model to overfitting.

\begin{table}
\caption{\label{tab:embedding-size}Impact feature embedding size. Best in bold.}
\begin{center}
\resizebox{0.49\textwidth}{!}{
\begin{tabular}{ccccc}
\hline
& Embedding size  &\makecell{$IDP\uparrow$}  & \makecell{$IDR\uparrow$} &\makecell{$IDF_1\uparrow$}   \\
\hline\hline

\multirow{4}{*}{\rotatebox{90}{\tiny Mask R-CNN}} & Baseline (2048) & \textbf{50.56} & \textbf{66.75} & \textbf{57.54}  \\
& 512 &  49.24 & 65.01 & 56.04 \\ 
& 1024 & 49.67 & 65.58 & 56.53\\

& 4096 & 49.70 & 65.62 & 56.56 \\
\hline
\multirow{4}{*}{{\rotatebox[origin=c]{90}{\tiny EfficientDet}}}& Baseline (2048) & \textbf{48.33} & \textbf{70.03} & \textbf{57.19}  \\
& 512 & 46.72 & 67.69  & 55.28 \\
& 1024 & 47.06  & 68.20  & 55.69 \\

& 4096 & 47.50 & 68.83 & 56.21  \\

\hline
\end{tabular}}

\end{center}
\end{table}

\begin{table}
\caption{\label{tab:Strategies}Impact of additional strategies. Best in bold. }
\begin{center}
\resizebox{0.49\textwidth}{!}{
\begin{tabular}{ccccc}
\hline
&Approach  &\makecell{$IDP\uparrow$}  & \makecell{$IDR\uparrow$} &\makecell{$IDF_1\uparrow$}   \\
\hline\hline
\multirow{6}{*}{\rotatebox{90}{ Mask R-CNN}} & Baseline & 50.56 & 66.75 & 57.54   \\
&+ Size filtering & 53.03 & 66.70 & 59.08    \\ 
&+ Blind Occlusion handling &  53.46 & 70.59 & 60.84 \\
&+ Reprojection-based handling  & 52.99 & 70.96 & 60.67   \\
&+ Blind Occlusion handling + Size filtering   & \textbf{54.99} & \textbf{69.17} & \textbf{61.27}\\
&+ Reprojection-based handl. + Size filtering & 54.06 & 69.02 & 60.63\\

\hline
\multirow{6}{*}{\rotatebox{90}{EfficientDet}} & Baseline &  48.33 & 70.03 & 57.19\\
&+ Size filtering   &  50.20   & 70.07 & 58.49\\ 
&+ Blind Occlusion handling &  53.18 & 77.05 &63.50\\
&+ Reprojection-based handling  & 51.53 & 75.67 & 61.31 \\
&+ Blind Occlusion handling + Size filtering   & \textbf{54.73} & \textbf{76.38} & \textbf{63.77}\\
&+ Reprojection-based handl. + Size filtering  & 53.34 & 75.50 & 62.52\\

\hline
\end{tabular}}

\end{center}
\end{table}

\textbf{Influence of additional strategies:} The additional strategies we have designed are divided in two branches: removing small detected objects that are not considered in the ground-truth, and occlusion handling.

To avoid the existing bias in the ground-truth towards distant cars that are not annotated, we performed a size filtering strategy by removing detections which area is under 0.10$\%$ of the total frame area. 

The blind occlusion handling and the reprojection-based occlusion handling strategies are detailed in Section \ref{sec:prop-temporal}.


Table \ref{tab:Strategies} shows the ablation results of these strategies. As expected, we can observe that the procedure of removing small detections increases the $IDP$ measure, using both object detectors, by 2.47 (1.87), while maintaining almost the same $IDR$. Since $IDP$ reacts to false positives, this seems to indicate that the size filtering removes those small detections we track, but are not annotated in the ground-truth. 

Both occlusion handling strategies improve the baseline tracking $IDR$ significantly, 3.81 (7.02) and 4.21 (5.64) respectively, and also the $IDP$ is being improved by 2.90 (4.85) and 2.43 (3.20). Contrary to expectations, the reprojection-based strategy is not overcoming the blind one. Another bias existing in the ground-truth could be the reason for explaining this, since occluded vehicles are not annotated.

When combining both occlusion handling strategies with size filtering, we achieve a higher precision than applying them separately, while recall is slightly narrowed. As in the previous comparison, these results suggest that the reprojection-based strategy does not provide improvements over the blind strategy due to the nature of the ground-truth. We consider using the baseline approach together with the blind occlusion handling and the size filtering strategies, a good trade-of between the $IDP$ and $IDR$.

\subsection{Comparison with the state-of-the-art}
Along this section, the proposed algorithm is compared with state-of-the-art approaches. Comparison is performed in the S02 scenario of the CityFlow benchmark, which is the only validation scenario with partially overlapping FOVs, as our method targets this scenario.

The proposed approaches in the literature devoted to vehicles MTMC tracking, listed in Table \ref{table:comparison}, have been already compared in the The 2019 AI City Challenge \cite{Naphade2019} jointly over the testing scenarios S02 and S05. However, as S05 consists of non-overlapping cameras, to ensure a fair comparison, we perform the evaluation only over S02. For this purpose, we ran the public available codes and we evaluated them following the CityFlow benchmark evaluation methodology, detailed in Section \ref{sec: evaluation-metrics}.

\begin{table}
\caption{\label{tab:results-SoA}Comparison with the state-of-the-art approaches. $\tau_{IoU}$ is the Intersection Over Union (IoU) evaluation threshold. The star (*) denotes that is an estimation. The extra 211 seconds are the duration of the video sequence under evaluation.}
\begin{center}
\resizebox{0.45\textwidth}{!}{
\begin{tabular}{cccccccc}
\hline
  &\makecell{$IDP\uparrow$}  & \makecell{$IDR\uparrow$} &\makecell{$IDF_1\uparrow$}  & $\tau_{IoU}$ & Processing & Latency (s) &  \makecell{Total \\Cost (min)} \\
\hline\hline
\multirow{2}{*}{UWIPL \cite{Hsu2019}}  & 70.21 & 92.61 &79.87  & 0.2 &   \multirow{8}{*}{Offline} & \multirow{2}{*}{3015* + 211} & \multirow{2}{*}{53.76*} \\
                        & 70.02 & 92.36 & 79.65 &0.5 &  &\\
 \multirow{2}{*}{ANU \cite{Hou2019}}   & 67.53  & 81.99 & 74.06 &0.2 & & \multirow{2}{*}{1159* + 211} & \multirow{2}{*}{22.83*}\\
                        & 66.42 & 80.64 & 72.85 & 0.5 &  &\\
 \multirow{2}{*}{BUPT \cite{He2019}}  & 78.23 & 63.69 & 70.22 & 0.2 &  &\multirow{2}{*}{1389* + 211} &\multirow{2}{*}{26.66*}\\
                        & 78.16 & 63.63 & 70.15 & 0.5 &  \\
 \multirow{2}{*}{NCCU \cite{Chang2019}}  & 48.91 & 43.35 & 45.97 & 0.2 & &  \multirow{2}{*}{2316* + 211}&  \multirow{2}{*}{42.11*}\\
                        & 24.36 & 21.59 & 22.89 & 0.5 & & \\
\multirow{2}{*}{Ours (EfficientDet)}  & 55.15 & 76.98 & 64.26 &  0.2 &  \multirow{4}{*}{Online} & \multirow{2}{*}{2.55} & \multirow{2}{*}{13.65}\\
                        & 54.73 & 76.38 & 63.77 & 0.5 &  & \\                        
\multirow{2}{*}{Ours (Mask-RCNN)}  & 57.23 & 71.99 & 63.76 &  0.2 &   & \multirow{2}{*}{2.29} & \multirow{2}{*}{12.71}\\
                        & 54.99 & 69.17 & 61.27 & 0.5 &  & \\

\hline
\end{tabular}}

\end{center}
\end{table}

Table \ref{tab:results-SoA} shows the evaluated performances in terms of $IDP$, $IDR$, $IDF_1$, latency and total computational time. The listed approaches can be divided by the processing mode in two groups: offline and online processing. As described in Section \ref{sec:soa}, to the best of our knowledge there is no previous proposal dealing with online MTMC vehicle tracking. For this reason, all the state-of-the-art methods that we evaluated are offline approaches. It is important to remark that, in Table \ref{tab:results-SoA}, the star denotes a partial and downward estimation. The codes for the complete systems are not publicly available, and only solutions based on precomputed intermediate results are accessible; hence, we can only compute the running time of the available modules. Therefore, the overall latency of the compared offline approaches is expected to be much higher than the results reported in Table \ref{tab:results-SoA}. Note that, the duration of the sequence under evaluation is also included in the latency since these offline approaches require access to results for the whole video to compute tracklets at each camera and then compute multi-camera tracks in a global way. As our proposal yields tracking results incrementally, from the beginning of the sequence, it can achieve a really low latency, in comparison with the others methods.

Regarding the quantitative measures of the tracking performance: $IDP$, $IDR$ and $IDF_1$, offline methods using constraining priors tailored to the target scenario clearly benefit from this extra information (see Table \ref{table:comparison}). In contrast to the rest of the state-of-the-art approaches, we are agnostic to the motion patterns of the vehicles (allowing to filter erroneous tracks), we do not perform any track post-processing (permitting to refining and unifying tracks and by this way reducing ID switches) and finally, we do not make use of manual annotations. On this basis, with an online approximation we perform really close to offline state-of-the-art approaches outperforming two of them in terms of Identification Recall.

Overall, our approach does not quite reach top performance in MTMC vehicle tracking, but its latency is three orders of magnitude smaller and the final computational cost is one order of magnitude faster, enabling a high performance operation on online mode with low-latency, that is a common requirement for many video-related applications, and also, in the generalization of the algorithms, avoiding hand-crafted strategies.

\section{Conclusion\label{sec:conclusions}}

Not relying on manual ad-hoc annotations, having no prior knowledge about the number of targets, and providing the best result in the shortest possible time are crucial requirements for a convenient and versatile algorithm. This paper presents, to the best of our knowledge, the first online MTMC vehicle tracking solution. Unlike previous approaches, the proposed approach continuously computes and updates the targets' state. 
We calculate clusters of detections of the same vehicle from different camera views applying a cross-camera clustering based on appearance and location. We train an appearance model to identify different views of the same vehicle leveraging homography matrices' information. Using information from the previous frame and a temporal estimation, we developed an occlusion handling strategy able to extrapolate accurate detections even if the target is occluded. Since the state estimation is continually updated, this strategy is useful even if the target is long-term occluded. 

This approach results in a low-latency MTMC vehicle tracking solution with quite promising results. Although performance is below its offline counterparts, the proposed one is a suitable solution for a real-world ITS technology.

\section*{Acknowledgement}
This work was partially supported by the Spanish Government (TEC2017-88169-R MobiNetVideo). 
We gratefully acknowledge the support of NVIDIA Corporation with the donation of the GPU used for the research of our group.

\ifCLASSOPTIONcaptionsoff
  \newpage
\fi



%

\bibliographystyle{IEEEtran}
\bibliography{main.bib}


%
\begin{IEEEbiography}[{\includegraphics[width=1in,height=1.25in,clip,keepaspectratio]{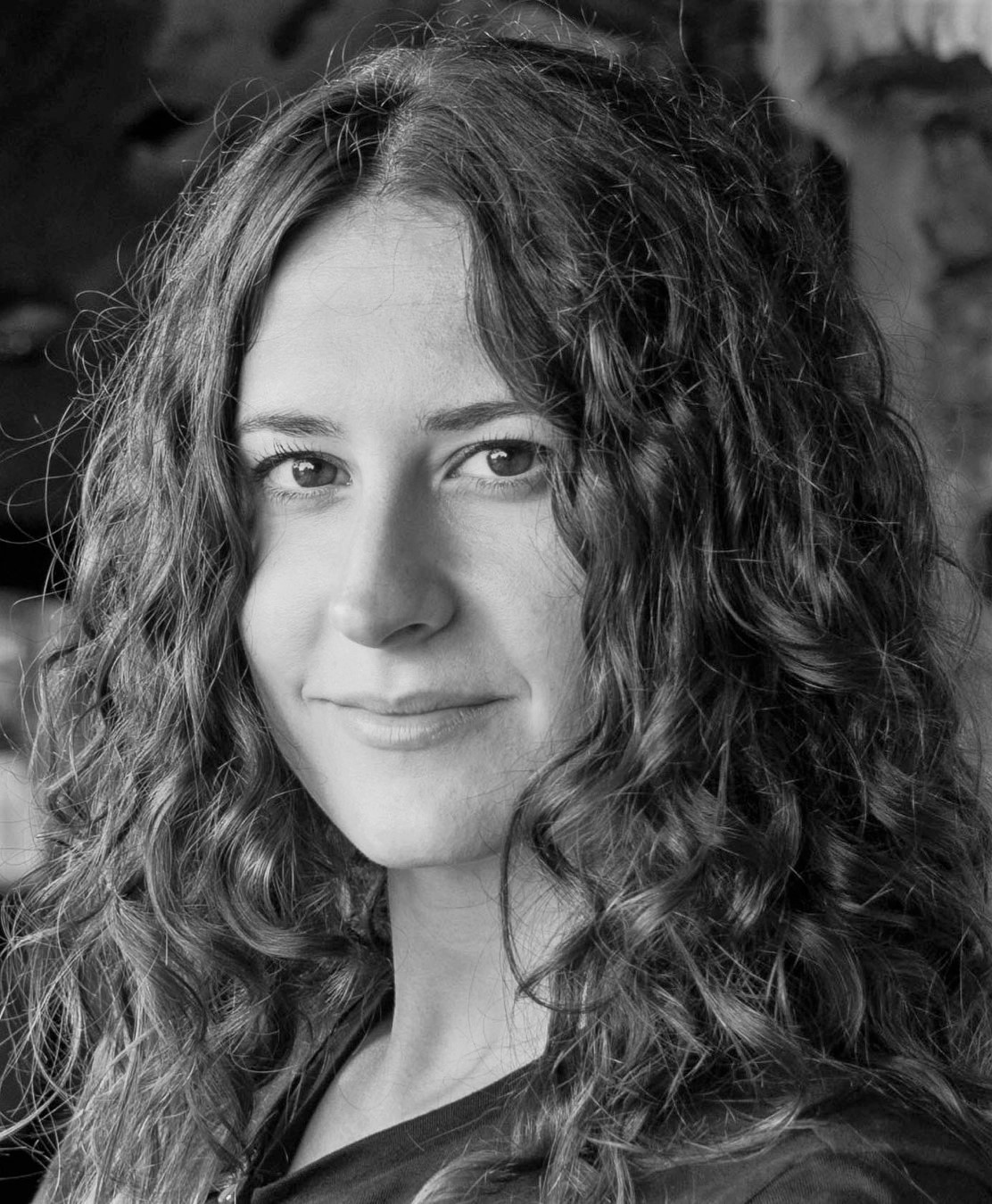}}]{Elena Luna García}

obtained a B.S degree in Telecommunications Engineering in 2015 at the Universidad Autónoma de Madrid (Spain). In 2017 she received the M.S. degrees belonging to the International Joint Master Program in Image Processing and Computer Vision (IPCV) at the PPCU in Budapest (Hungary), the University of Bordeaux (France) and the UAM (Spain). She is currently pursuing the Ph.D. degree with the Video Processing and Understanding Lab (VPU-Lab) at the UAM (Spain).

\end{IEEEbiography}

\begin{IEEEbiography}[{\includegraphics[width=1in,height=1.25in,clip,keepaspectratio]{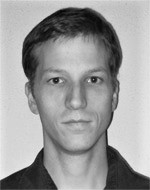}}]{Juan C. SanMiguel}
received the Ph.D. degree
in computer science and telecommunication from
University Autonoma of Madrid, Madrid, Spain,
in 2011. He was a Post-Doctoral Researcher with
Queen Mary University of London, London, U.K.,
from 2013 to 2014, under a Marie Curie IAPP
Fellowship. He is currently Associate Professor at
University Autónoma of Madrid and Researcher
with the Video Processing and Understanding Laboratory. His research interests include computer vision
with a focus on online performance evaluation and
multicamera activity understanding for video segmentation and tracking. He
has authored over 40 journal and conference papers.
\end{IEEEbiography}

\begin{IEEEbiography}[{\includegraphics[width=1in,height=1.25in,clip,keepaspectratio]{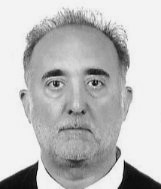}}]{ José M. Martínez}received the Ph.D. degree in
computer science and telecommunication from the
Universidad Politécnica de Madrid, Madrid, Spain,
in 1998. He is currently a Full Professor with the
Escuela Politécnica Superior, Universidad Autónoma
de Madrid, Madrid. He has acted as an auditor and
a reviewer for the EC for projects of the frameworks
program for research in Information Society and
Technology (IST). He is the author or coauthor of
more than 100 papers in international journals and
conferences and a coauthor of the first book about
the MPEG-7 standard published in 2002. His professional interests cover
different aspects of advanced video surveillance systems and multimedia
information systems.

\end{IEEEbiography}

\begin{IEEEbiography}[{\includegraphics[width=1in,height=1.25in,clip,keepaspectratio]{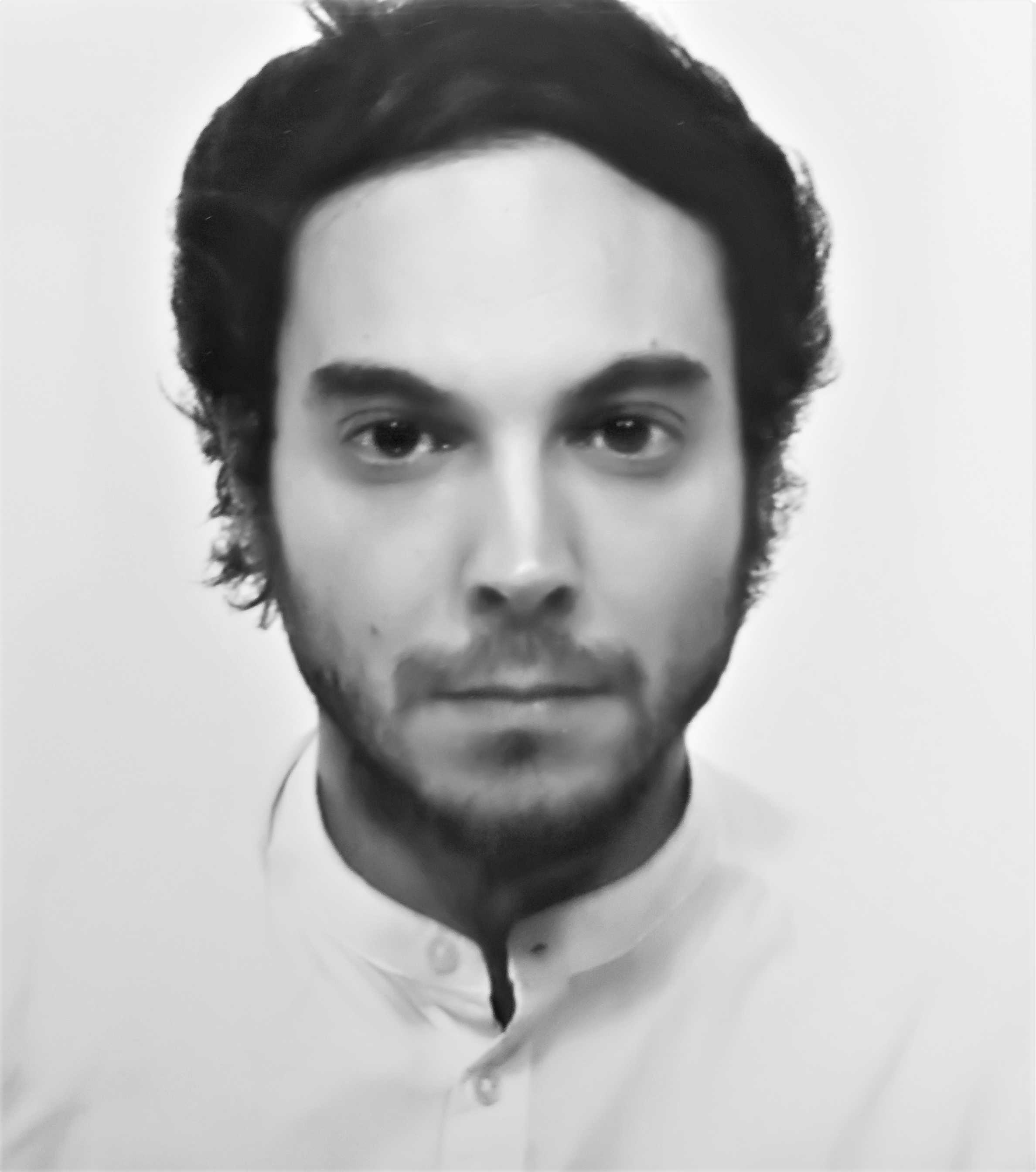}}]{ Marcos Escudero-Viñolo}received the M.Sc. degree in Telecommunications Engineering (Feb. 2008), the M.Phil. Degree in Multimedia Signal Processing (Nov. 2009) and the Ph.D. degree in Telecommunications Engineering and Computer Science (Jan. 2016) from the Universidad Autónoma de Madrid (UAM), Spain. He is currently a researcher and Teaching Assistant at UAM. Since 2007, he has been researching with the Video Processing and Understanding Laboratory (VPU-Lab). His research has been funded by European (i.e. IST-FP6-027685 MESH) and National projects (i.e. CENIT-VISION 2007-1007)—both from the public sector and private companies—, and by the Spanish Government through its FPU program (2010-2014). He has been a three-month research visitor at Queen Mary University (May-July 2013). His research is mainly focused on the area of image and video analysis and understanding, both in the compress and uncompressed domains, analyzing colour and depth data and targeting to develop automatic video content translators, with a special focus on region-driven analysis. He serves as a regular reviewer for international conferences (i.e. ICIP) and journals (i.e. IEEE Tr. on Image Processing, Pattern Recognition Letters, Multimedia Tools and Applications

\end{IEEEbiography}





\end{document}